\pdfoutput=1

\documentclass[11pt]{article}

\usepackage[]{EMNLP2023}

\usepackage{times}
\usepackage{latexsym}

\usepackage[T1]{fontenc}

\usepackage[utf8]{inputenc}

\usepackage{microtype}

\usepackage{inconsolata}

\usepackage{hyperref}
\usepackage{graphicx}
\usepackage{tabularx}
\usepackage{multirow}

\title{Text Style Transfer Evaluation Using Large Language Models}

\author{Phil Ostheimer, Mayank Nagda, Marius Kloft, Sophie Fellenz \\
        RPTU Kaiserslautern-Landau \\
        surname@cs.uni-kl.de\\}

\begin{document}
\maketitle
\begin{abstract}
Evaluating Text Style Transfer (TST) is a complex task due to its multifaceted nature. The quality of the generated text is measured based on challenging factors, such as style transfer accuracy, content preservation, and overall fluency. While human evaluation is considered to be the gold standard in TST assessment, it is costly and often hard to reproduce. Therefore, automated metrics are prevalent in these domains. Nevertheless, it remains unclear whether these automated metrics correlate with human evaluations.
Recent strides in Large Language Models (LLMs) have showcased their capacity to match and even exceed average human performance across diverse, unseen tasks. This suggests that LLMs could be a feasible alternative to human evaluation and other automated metrics in TST evaluation.
We compare the results of different LLMs in TST using multiple input prompts. Our findings highlight a strong correlation between (even zero-shot) prompting and human evaluation, showing that LLMs often outperform traditional automated metrics. Furthermore, we introduce the concept of prompt ensembling, demonstrating its ability to enhance the robustness of TST evaluation.
This research contributes to the ongoing evaluation of LLMs in diverse tasks, offering insights into successful outcomes and areas of limitation.
\end{abstract}
\section{Introduction}
\begin{figure}[ht!]
\centering
\includegraphics[width=0.48\textwidth]{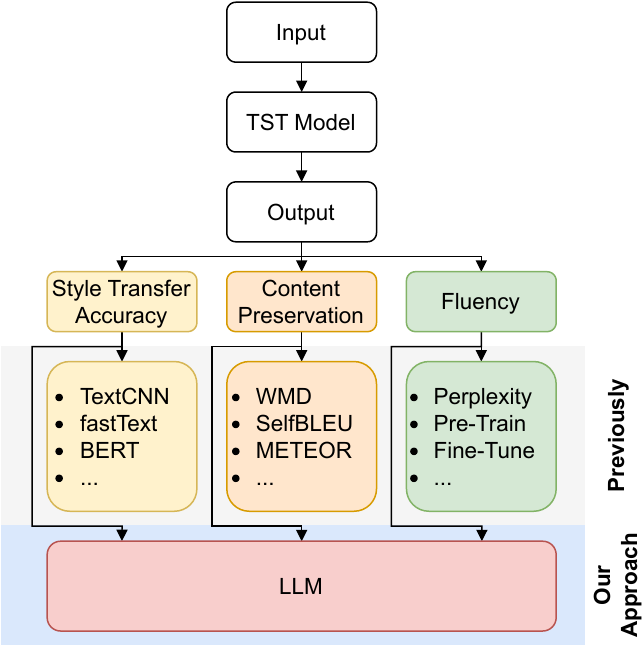}
\caption{Shown is our approach to TST evaluation by replacing the multitude of (non-)validated metrics by LLM evaluation for a unified TST evaluation. In our approach, an LLM measures all three aspects of TST evaluation: style transfer accuracy, content preservation, and fluency.}
\label{fig:approach}
\end{figure}

Text Style Transfer (TST) aims to change a text's style while retaining its content \cite{Jin:2022}. Examples of different TST tasks include sentiment transfer \cite{Shen:2017}, politeness transfer \cite{Niu:2018}, and formality transfer \cite{Rao:2018}, to name just a few. TST is usually evaluated in terms of multiple aspects, foremost style transfer accuracy, content preservation, and fluency of the text \cite{Mir:2019}. Style transfer accuracy assesses how closely the generated style matches the target style, content preservation evaluates how well the original content has been preserved, and fluency measures the overall naturalness of the text. However, separating these aspects proves to be a challenging task \cite{Jafaritazehjani:2020}, given the vast diversity of text styles. For example, a text's sentiment can be considered both a style aspect or an integral part of the content. Due to these challenges, it is essential to invest additional research effort into TST evaluation. This includes comparing the performance of various TST methods to measure the progress made within the field.

Human evaluation is widely regarded as the most reliable evaluation method in many NLP tasks, including natural language generation and TST \cite{Briakou:2021a}. However, human evaluation does have limitations, particularly when the evaluators are not domain experts in the specific task evaluated \cite{Clark:2020}. Additionally, there are severe concerns regarding underspecification, availability, reliability, lack of standardization, and reproducibility of human evaluation \citep{Briakou:2021a}. Lastly, human evaluation can be costly and time-consuming. As a result, many studies rely on automated metrics as a substitute for human evaluation due to the limitations and challenges associated with the latter.

Despite certain advancements \cite{Mir:2019,Pang:2019, Briakou:2021b, Yamshchikov:2021}, the extent to which automated metrics can serve as reliable substitutes for human evaluation remains uncertain. Numerous TST publications employ non-validated automated metrics or metrics that are known to be suboptimal \cite{Ostheimer:2023}. Additionally, the abundance of available automated TST evaluation metrics raises concerns regarding the standardization of the evaluation procedure \cite{Ostheimer:2023}.

We propose employing Large Language Models (LLMs) as evaluators for TST. LLMs have demonstrated remarkable few-shot and zero-shot performance in diverse NLP tasks \cite{Brown:2020,Liu:2023} and have recently proven effective as NLP task evaluators \cite{Cheng:2023}. Our study aims to explore the potential of LLMs in replacing automated metrics across all three aspects of TST evaluation (see Fig. \ref{fig:approach}), thereby standardizing the evaluation practices. Our contributions can be outlined as follows:
\begin{enumerate}
    \item We propose using LLMs as a unified evaluation method for TST, covering all three essential aspects: style transfer accuracy, content preservation, and fluency.
    \item We experiment with multiple LLMs using zero-shot prompting. The results indicate that, across various settings, LLMs correlate better with human evaluations than previous automated metrics of TST quality.
    \item We demonstrate that the robustness of LLM evaluation can be improved by ensembling multiple prompts, mitigating the need for extensive prompt engineering.
\end{enumerate}

\section{Related Work}
In this section, we first discuss existing work on automated TST evaluation. This includes previous standardization and validation efforts and existing metrics. In the second part, we give an overview of different LLMs. In the last part, we introduce related work on using LLMs for unseen tasks, including evaluation.

\subsection{Standardization and Validation of Automated TST Evaluation}
Several previous studies aim to standardize and validate the automated TST evaluation. Notably, \citet{Mir:2019} make significant contributions by developing enhanced automated evaluation metrics to assess style transfer accuracy, content preservation, and fluency. Additionally, \citet{Pang:2019} explore the correlations between automated metrics and human evaluations for all three aspects. Furthermore, \citet{Yamshchikov:2021} conduct a comprehensive large-scale study to identify the most effective automated metric specifically for content preservation. It is worth mentioning that, although our research focuses solely on TST in English, \citet{Briakou:2021b} investigates various automated metrics for style transfer accuracy, content preservation, and fluency in a multilingual setting, aiming to identify those with the highest correlation to human evaluations.

However, it is important to note, as highlighted by \citet{Ostheimer:2023}, that TST evaluation still lacks standardization. There exists a wide range of different automated metrics for each evaluation aspect. Moreover, numerous publications employ either non-validated automated metrics or metrics that do not exhibit the highest correlation with human evaluations.

\subsubsection{Style Transfer Accuracy}
To measure style transfer accuracy, the prevalent method in automated evaluation involves employing a sentence-level style classifier, as established in previous works \cite{Mir:2019, Pang:2019, Ostheimer:2023}. Notably, automated metrics such as TextCNN \cite{Kim:2014}, fastText \cite{Joulin:2017}, and BERT \cite{Devlin:2019} fine-tuned for style classification have gained popularity for this purpose \cite{Ostheimer:2023}.

\subsubsection{Content Preservation}
When evaluating content preservation, it is customary to draw on count-based metrics used in the machine translation domain, such as BLEU \cite{Papineni:2002} and METEOR \cite{Banerjee:2005}. Additionally, embedding-based metrics, including the embedding average \cite{Mir:2019}, greedy matching \cite{Rus:2012}, vector extrema \cite{Forgues:2014}, and word mover's distance (WMD) \cite{Kusner:2015}, are commonly employed.

\subsubsection{Fluency}
Fluency in TST is often evaluated by calculating the perplexity of a pre-trained or fine-tuned language model \cite{Mir:2019, Pang:2019}. However, it is important to note that there exists a wide variety of language model architectures and training methods \cite{Ostheimer:2023}.

In previous studies \cite{Mir:2019, Pang:2019}, which explore the suitability of language models for TST evaluation, the focus is primarily on measuring fluency. Perplexity is used as the metric for assessing fluency, yielding mixed results. While \citet{Mir:2019} report a limited correlation, \citet{Pang:2019} find a high correlation. In contrast, our approach involves using language models to evaluate all three TST aspects and demonstrate their validity.

\subsection{Large Language Models}
Large Language Models (LLMs) are characterized by their extensive parameter count, often reaching billions. These models are typically pre-trained on vast datasets. Prominent examples of LLMs include GPT3 \cite{Brown:2020}, OPT \cite{Zhang:2022}, BLOOM \cite{Scao:2022}, and more recently Falcon \cite{Almazrouei:2023}, and Llama2 \cite{Touvron:2023}. What sets LLMs apart is their ability to perform valuable tasks even without fine-tuning, showcasing their zero- and few-shot capabilities \cite{Liu:2023}. However, it has been demonstrated by \citet{Ouyang:2022} that larger LLMs are not inherently superior at understanding user intent. Nevertheless, aligning these models with user intent through Supervised Fine-Tuning (SFT) and Reinforcement Learning from Human Feedback (RLHF) can significantly enhance their performance.

The emergence of LLMs introduces a new paradigm known as prompting \cite{Liu:2023}. Prompting involves solving prediction tasks without the need for fine-tuning or additional training. Solving an unseen task involves modifying an input $x$ using a template to create a textual string $x'$ where $x'$ contains an empty slot to be filled by the LLM. Previous work \cite{Liu:2023} distinguishes between cloze prompts, where the empty slot can be anywhere in the prompt, and prefix prompts, where the to-be-filled slot is at the end. The LLM is then employed to fill in the prompt, resulting in a final string $\hat{x}$, subsequently parsed to extract the desired answer. 

There is a long history of combining multiple systems to improve the performance \cite{Wolpert:1992,Zhou:2002} of machine learning systems. For LLMs, multi-prompt learning combines multiple prompts to make prompting more effective. A notable approach within multi-prompt learning is prompt ensembling, a technique where answers from multiple prompts can be averaged. Prompt ensembling leverages the benefits of using multiple prompts while mitigating the challenges of prompt engineering, ultimately leading to potentially more robust downstream performance \cite{Liu:2023}.

\subsection{LLM Evaluation}
In their research, \citet{Cheng:2023} introduce the term ``LLM evaluation'' to refer to the evaluation of NLP tasks using LLMs. They specifically focus on evaluating open-ended text generation and adversarial attacks across various evaluation aspects. Their study reveals that LLMs can distinguish between human-written and machine-generated text and they report varying correlations for different evaluation aspects, ranging from weak to strong.

\citet{Cheng:2023} experiment with several RLHF/SFT-aligned LLMs showing their suitability, even in the zero-shot setting. In contrast, our study compares RLHF/SFT-improved LLMs (specifically InstructGPT, Falcon, and Llama2) to non-RLHF/SFT-improved LLMs (OPT, BLOOM, GPT3, and also Falcon and Llama2 in their pre-trained versions).

Furthermore, preliminary work conducted by \citet{Gilardi:2023} and \citet{Huang:2023} suggests that LLM evaluation outperforms human evaluation in tasks such as text classification and explanation of implicit hate speech, showcasing the superior performance of LLMs in these domains. However, so far, no one has used LLMs in order to evaluate TST.

\section{Method}
\begin{figure*}[ht!]
\centering
\includegraphics[width=\textwidth]{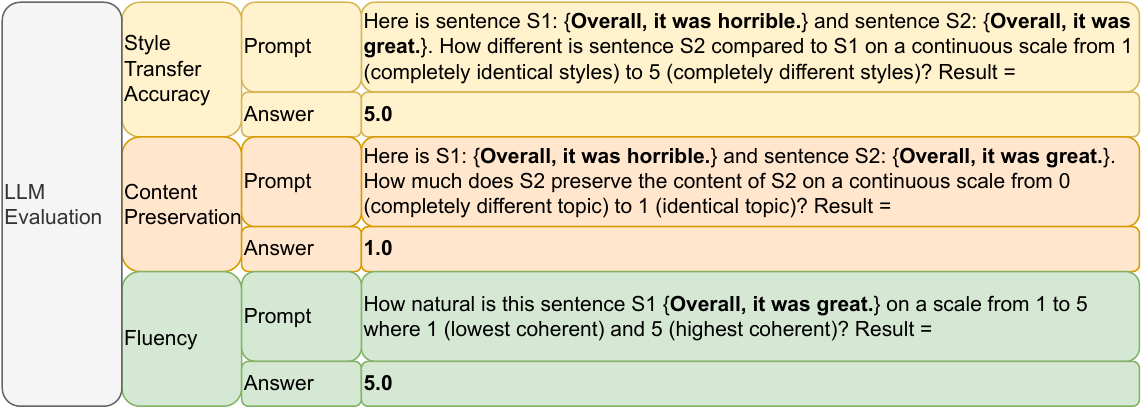}
\caption{Shown is our method for TST evaluation using LLMs. We present one prompt for one example input and output and its parsed answer, a score limited by the given range. The shown prompts are the ones exhibiting the highest correlation with human evaluations.}
\label{fig:method}
\end{figure*}

\paragraph{Unifying TST Evaluation} Our approach is illustrated in Figure \ref{fig:approach}, where we propose a novel methodology to replace the extensive array of existing automated metrics by leveraging LLMs. Unlike previous approaches that rely on a language model's perplexity for fluency assessment, our method employs LLMs to evaluate all three aspects comprehensively. However, we restrict our analysis to the numerical scores provided by the LLM limited to be within the given score scale. This means that any score returned by the model that falls outside the allowed range is ignored and we exclude the respective data point from the reported results. The LLM used in our approach can vary in size and may have undergone different pre-training and fine-tuning techniques, as detailed in Section \ref{sec:llms}.

\paragraph{Prompting} In Figure \ref{fig:method}, we show one example prompt per evaluation aspect (the complete list of prompts can be found in Appendix \ref{sec:prompts}) . The prompts are filled with an input/output example. In our case, the answer is a score parsed using a simple regular expression from the suffix that the LLM added to the prompt. In total, we experiment with 11 prompts per aspect with different scales, including continuous scales from 0 to 1 and continuous and discrete scales from 1 to 5 (similar to a Likert scale). For the design of the prompts, we experiment with zero-shot prompts that directly ask for the evaluation of a certain aspect of TST evaluation with no further examples given. We take inspiration from previous human evaluations and their questionnaire design \cite{Briakou:2021a} for the design of the prompts. The prompts are designed as prefix prompts such that the answer (in our case, a numerical score) can be easily parsed using regular expressions.

\paragraph{Prompt Ensembling} To increase and demonstrate the robustness against different prompt formulations of our approach, we employ an ensemble of multiple prompts. We normalize the scores and afterward apply a uniform averaging across the prompts per aspect. This ensemble approach allows us to mitigate the impact of individual prompt variations and provide a more reliable assessment.

\section{Experimental Setup}
\subsection{TST Models}
To evaluate the TST evaluation capabilities of LLMs, we consider three well-known TST models for which human evaluation results are publicly available. We namely experiment with outputs from the Cross-Aligned Autoencoder (CAAE) \cite{Shen:2017}, Adversarially Regularized Autoencoder (ARAE) \cite{Zhao:2018}, and delete-and-retrieve (DAR) \cite{Li:2018}. We evaluate the TST models' output with the LLMs and compute the correlations to human evaluations to show their performance compared to existing automated metrics. For each model, we evaluate the available human-annotated sentences, comprising an equal number of positive and negative examples, from the test set (244), totaling 732 sentences provided by \citet{Mir:2019} for the Yelp dataset \cite{Shen:2017}.

\subsection{LLMs}
\label{sec:llms}
We experiment with six LLMs: OPT \cite{Zhang:2022}, BLOOM \cite{Scao:2022}, GPT3 \cite{Brown:2020}, InstructGPT \cite{Ouyang:2022}, Falcon \cite{Almazrouei:2023}, and Llama2 \cite{Touvron:2023}.  The LLMs can be grouped into two groups. Pre-trained LLMs: OPT, BLOOM, GPT3, Falcon, and Llama2 (pre-trained with the ordinary autoregressive language modeling objective) and the LLMs fine-tuned (with RLHF/SFT) to follow the user intent:  InstructGPT, Falcon, and Llama2 (Falcon and Llama2 were used in their pre-trained and fine-tuned versions). To access GPT3 (in the version \textit{davinci} with 175 billion parameters) and InstructGPT (in the version \textit{text-davinci-003} with 175 billion parameters), we use the API provided by OpenAI \footnote{\url{https://openai.com/api/}}, as for OPT, BLOOM, Falcon, and Llama2 we deploy the pre-trained models on our own hardware using Alpa \cite{Zheng:2022}. 

To investigate the impact of different Language Model sizes, we use the OPT \cite{Zhang:2022} family of models, as well as Falcon \cite{Almazrouei:2023} and Llama2 \cite{Touvron:2023}. The OPT models have demonstrated performance comparable to GPT2 \cite{Radford:2019} and GPT3 \cite{Brown:2020}. An advantage of the OPT models is that most are freely available for use (except for the largest OPT-175b, which requires a request for access). The OPT models we employ have sizes of 125m, 350m, 1.3b, 2.7b, 6.7b, 13b, 30b, 66b, and 175b parameters. We use the 7b and 40b models for Falcon in the "normal" and "instruct" version. For Llama2, we run the models with 7b, 13b, and 70b parameters in the ``normal'' and ``chat'' versions. 

\section{Correlations of LLM Evaluations with Human Evaluations}
\label{sec:correlations}
In order to evaluate the effectiveness of LLMs for the task of TST evaluation, we measure the correlation between the evaluations generated by LLMs and the corresponding human evaluations for each aspect. In this section and the subsequent sections, our primary focus is on the largest LLMs that were fine-tuned to follow instructions, namely InstructGPT \cite{Ouyang:2022}, Falcon \cite{Almazrouei:2023} ("instruct" version with 7b and 40b parameters), and Llama2 \cite{Touvron:2023} ("chat" version with 7b, 13b, and 70b parameters). We prioritize these larger, fine-tuned models due to the limited reliability observed in the evaluations conducted with smaller, non-fine-tuned models in our setting with zero-shot prompts (refer to Appendix \ref{sec:smaller_llms} for further information).

\subsection{Effect of Ensembling}
\label{sec:effect_of_ensembling}

\begin{figure}[ht!]
    \centering
    \includegraphics[width=0.47\textwidth]{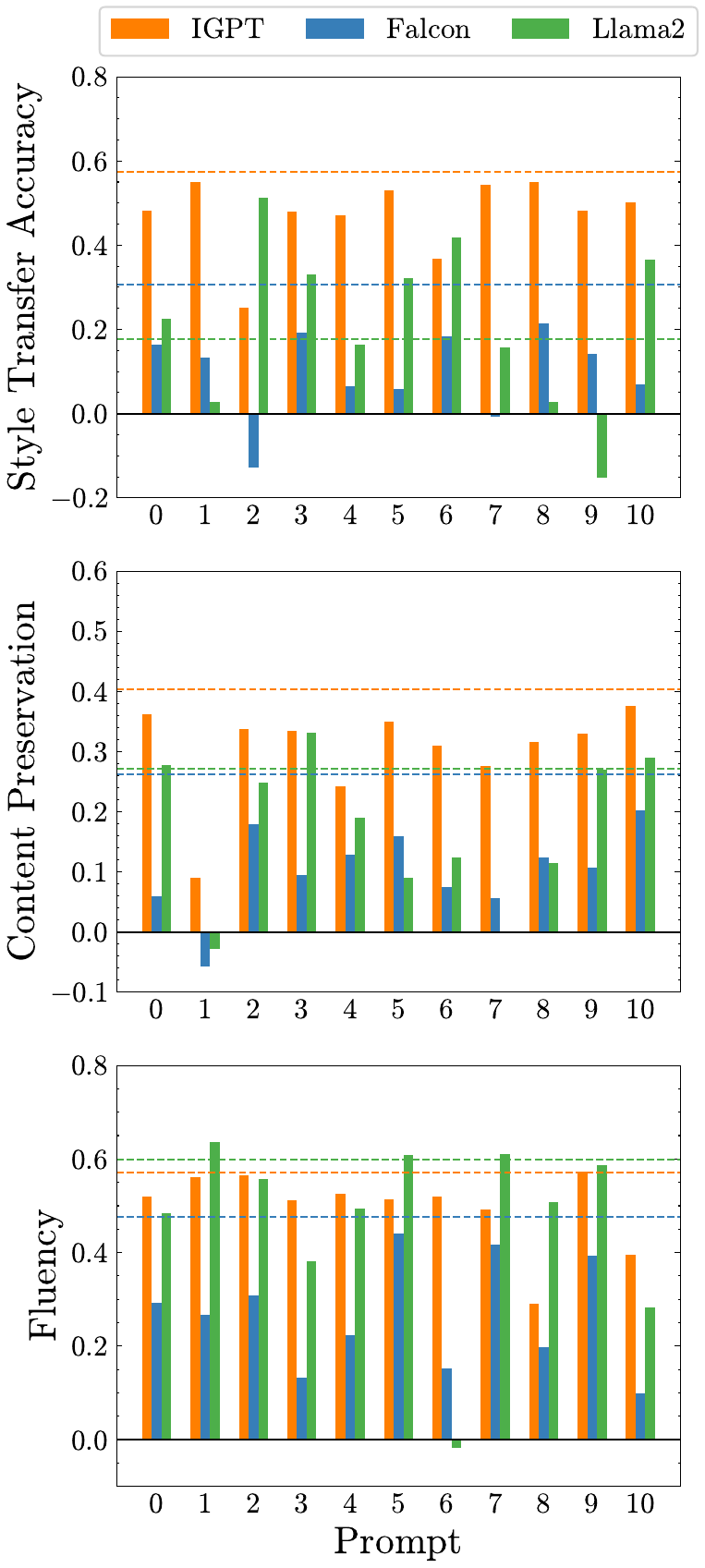}
    \caption{Shown is the Spearman rank correlation of each prompt for measuring style transfer accuracy (top), content preservation (middle), and fluency (bottom) with human evaluations for InstructGPT (IGPT with 175b parameters),  Falcon (``instruct'' with 40b parameters), and Llama2 (``chat'' with 70b parameters). The horizontal dashed lines indicate the correlation of the ensembling of all prompts.}
    \label{fig:llm_correlations}
\end{figure}

This section shows how ensembling improves the robustness of our zero-shot prompting approach for TST evaluation. Figure \ref{fig:llm_correlations} shows style transfer accuracy on top, content preservation in the middle, and fluency on the bottom. The correlations between the returned scores from each prompt and the human evaluations are represented as bars, while the correlation of the ensembled prompts is depicted as a horizontal dashed line.

\paragraph{Style Transfer Accuracy} InstructGPT has the highest correlations for individual prompts. InstructGPT's and Falcon's ensembled prompts' correlation surpasses that of the individual prompts. However, for Llama2, we observe greater variations in the correlations for particular prompts. The divergence between InstructGPT, Falcon, and Llama2 for prompt 2 can be attributed to slight variations in sentence placeholders, where ``S1'' and ``S2'' were replaced with ``A'' and ``B'' respectively. Especially Falcon returns for this prompt mostly unparsable answers. Prompt 9 seems to be challenging because it explicitly asks for a rating on a continuous scale. 

\paragraph{Content Preservation} For all prompts, InstructGPT shows the highest correlations. InstructGPT's and Falcon's ensembled prompts correlate more than the individual prompts. For Llama2, we observe less variance in the correlations for the individual prompts than for style transfer accuracy. However, the ensembled correlation is still on par or better than most of the individual prompts, indicating that also for content preservation, ensembling makes the evaluations more robust. The weaker performance of prompt 1 can be attributed to slight changes in the question formulation (from a quantitative to a qualitative question) and the inversion of the scale, resulting in many unparsable filled prompts.

\paragraph{Fluency} InstructGPT has, for most prompts, the highest correlations, closely followed by Llama2. InstructGPT's ensembled prompts perform on par or even better than the individual prompts. The correlation of ensembled prompts surpasses Llama2's majority of correlations of individual prompts. For Falcon, the ensembled prompts have the highest correlation. The low correlation of prompt 8 for InstructGPT can be attributed to the fact that here, we do not refer to fluency or naturalness but only to coherence as a synonym. Prompt 6 mentions grammar instead of naturalness or fluency and seems to be especially challenging for Falcon. Also, prompt 10 directly inputs the to-be-evaluated sentence without the task description as a prefix.

\subsection{Ensembled Prompts vs. Other Automated Evaluations}
\begin{table}[ht!]
\centering
\begin{tabular}{lrrrr}
\hline
\multicolumn{5}{c}{Style Transfer Accuracy} \\
\hline
\hline
& ARAE & CAAE & DAR & All\\
\hline
fastText & 0.498 & \textbf{0.550} & 0.332 & 0.473\\
TextCNN & 0.512 & 0.525 & 0.331 & 0.458\\
BERT & 0.513 & 0.559 & 0.408 & 0.497 \\
\hline
IGPT & \textbf{0.618} & 0.543 & \textbf{0.584} & \textbf{0.574}\\
Fal-7b & \emph{-0.027} & -0.219 & \emph{-0.118} & -0.131 \\
Fal-40b & 0.206 & 0.389 & 0.313 & 0.307 \\
Lla-7b & \emph{0.091} & -0.128 & \emph{-0.064} & \emph{-0.039} \\
Lla-13b & \emph{0.103} & \emph{0.018} & \emph{0.106} & \emph{0.067} \\
Lla-70b & 0.347 & \emph{0.075} & \emph{0.077} & 0.178 \\
\hline
\multicolumn{5}{c}{Content Preservation} \\
\hline
\hline
 & ARAE & CAAE & DAR & All \\
\hline
BLEU & 0.197 & 0.451 & 0.403 & 0.339 \\ 
METEOR & 0.247 & \textbf{0.659} & \textbf{0.425} & \textbf{0.420} \\
EmbAvg & \emph{0.087} & 0.500 & 0.269 & 0.273 \\
GrMatch & 0.203 & 0.592 & 0.377 & 0.358 \\
VecExtr & 0.189 & 0.503 & 0.390 & 0.328 \\
WMD & \textbf{0.240} & 0.615 & 0.361 & 0.377 \\
\hline
IGPT & 0.191 & 0.656 & 0.345 & 0.404\\
Fal-7b & \emph{-0.022} & \emph{0.050} & \emph{-0.016} & \emph{0.012} \\
Fal-40b & 0.167 & 0.386 & 0.240 & 0.262 \\
Lla-7b & \emph{-0.035} & \emph{0.052} & \emph{0.120} & \emph{0.061} \\
Lla-13b & \emph{-0.099} & \emph{-0.064} & 0.157 & \emph{0.040} \\
Lla-70b & \emph{0.104} & 0.484 & 0.198 & 0.271 \\
\hline
\multicolumn{5}{c}{Fluency} \\
\hline
\hline
 & ARAE & CAAE & DAR & All \\
\hline
PPL PT & \emph{0.076} & \emph{0.044} & 0.418 & 0.171 \\ 
PPL FT & 0.135 & \emph{0.120} &  0.411 & 0.232 \\
\hline
IGPT & 0.518 & \textbf{0.560} & \textbf{0.603} & 0.571\\
Fal-7b & \emph{-0.057} & \emph{0.075} & \emph{-0.081} & \emph{-0.010} \\
Fal-40b & 0.436 & 0.452 & 0.491 & 0.476 \\
Lla-7b & 0.172 & 0.143 & 0.311 & 0.216\\
Lla-13b & 0.184 & 0.200 & 0.459 & 0.290 \\
Lla-70b & \textbf{0.539} & 0.551 & 0.602 & \textbf{0.599}\\
\hline
\end{tabular}
\caption{Shown are the Spearman rank correlations for style transfer accuracy (top), content preservation (middle), and fluency (bottom) between human evaluations and the mentioned automated metrics, including InstructGPT (IGPT), Falcon (Fal), and Llama2 (Lla). All \emph{italic correlations} have p>0.05.}
\label{tab:automated_evaluation_correlations}
\end{table}
Table \ref{tab:automated_evaluation_correlations} compares automated state-of-the-art TST evaluation measures to our approach with LLM evaluation. Table \ref{tab:automated_evaluation_correlations} shows style transfer accuracy on top, content preservation in the middle, and fluency on the bottom.

\paragraph{Style Transfer Accuracy} We compare our approach to fastText \cite{Joulin:2017}, TextCNN \cite{Kim:2014}, and a BERT-based \cite{Devlin:2019} classifier fine-tuned for style classification. As can be seen, InstructGPT (IGPT) has a higher correlation with human evaluations for each TST model, except CAAE, where it is slightly worse than fastText \cite{Joulin:2017}. However, if looking at the combined model outputs, InstructGPT has the highest correlation to human evaluations compared to the other automated style transfer accuracy metrics. For Falcon and Llama2, we can observe that the smaller models with 7b or 13b parameters show relatively low or statistically insignificant correlations. The biggest models with 40b and 70b parameters show smaller correlations than InstructGPT.

\paragraph{Content Preservation} As count-based metrics, we report the (Self-)BLEU score \cite{Papineni:2002,Briakou:2021b} between the input and output and METEOR \cite{Banerjee:2005}. Among the embedding-based metrics, we report greedy matching \cite{Rus:2012}, vector extrema \cite{Forgues:2014}, and word mover's distance (WMD) \cite{Kusner:2015}. InstructGPT has slightly lower correlations with human evaluations than WMD and METEOR for ARAE and DAR, respectively. CAAE evaluations with METEOR are very close to InstructGPT. However, combining the outputs from all three TST models, we see a similar correlation for InstructGPT with the best automated metric, METEOR. Falcon's and Llama2's smaller models with 7b or 13b parameters show relatively low or statistically insignificant correlations, while the biggest models with 40b and 70b parameters show smaller correlations than InstructGPT.

\paragraph{Fluency} We compare our approach to a pre-trained and a fine-tuned (on the Yelp dataset) GPT2 \cite{Radford:2019} measuring perplexity (PPL). As can be seen, InstructGPT exposes the highest correlations with human evaluations for CAAE and DAR. At the same time, Llama2 shows the highest correlation for ARAE and for combining ARAE's, CAAE's, and DAR's output. However, for fluency, the biggest models, InstructGPT, Falcon with 40b parameters, and Llama2 with 70b parameters show considerably higher correlations than measuring the perplexity with GPT2. Also, Llama2's smaller models with 7b and 13b parameters noticeably correlate with human evaluations, and only Falcon's smallest model with 7b parameters shows insignificant correlations.

\section{LLM Limitations and Failure Modes}
In this section, we summarize several qualitative limitations of our approach.

\subsection{Unparsable Filled Prompts}
\label{sec:unparsable_llm_answers}
\begin{table}[htpb!] 
\centering
\begin{tabular}{lrrr}
\hline
 & STA & CP & F \\
\hline
\hline
IGPT & 0.0\% & 0.0\% & 0.0\% \\ 
Falcon-7b & 0.0\% & 0.2\% & 0.1\%\\
Falcon-40b & 7.9\% & 9.4\% & 10.2\% \\
Llama2-7b & 24.4\% & 40.5\% & 1.9\% \\
Llama2-13b & 14.7\% & 19.2\% & 1.0\%\\
Llama2-70b & 30.9\% & 28.6\% & 1.1\%\\
\hline
\end{tabular}
\caption{Shown is the proportion of filled prompts for the three fine-tuned LLMs evaluation models InstructGPT (IGPT), Falcon, and Llama2 in different model sizes where the answer is not parsable to return a score for the aspects of style transfer accuracy (STA), content preservation (CP), and fluency (F). InstructGPT is the most reliable.}
\label{tab:unparsable_answers}
\end{table}

\begin{table*}[htpb!]
\centering
\begin{tabular}{p{2cm}p{2.5cm}p{10cm}}
\hline
Prompt & Input & Here is sentence S1: \{honestly they were down right rude .\} and sentence S2: \{honestly they were down right , friendly and fast !\}. How different is sentence S2 compared to S1 on a scale from 1 (identical styles) to 5 (completely different
styles)? Result =\\
\hline
\hline
\multirow{2}{2cm}{Filled Prompt} & IGPT & [...]5 (completely different styles)\\
& Falcon-7b & [...]" 3 S1 is 4 (more similar styles) and S2 is 5 (completely different styles)"\\
& Falcon-40b & [...]" 3 The difference between sentence S1 and S2 is a score of 3 out of 5."\\
& Llama2-7b & [...]" To compare the styles of sentences S1 and S2, we need to analyze their grammatical structures, word choices, and tone. Here's a detailed analysis: 1. Grammar: Sentence S1: * Subject: They were * Verb: were * Ad"\\
& Llama2-13b & [...]" Sure, I can help you with that! Sentence S1: \{honestly they were down right rude .\} Sentence S2: {honestly they were down right, friendly and fast !} The difference between these two sentences is a 5"\\
& Llama2-70b & [...]" The two sentences have different styles, but they share some similarities. S1 has a negative tone, using the word ""rude"" to describe someone's behavior. The sentence structure is also simple, with a single clause expressing a complete thought. S2, on the other hand" \\
\hline
\end{tabular}
\caption{Shown are example LLM evaluations for DAR output. We omit the repetition of the prefix prompt for the filled prompts and denote it by [...]. Line breaks are ignored to fit table.}
\label{tab:llm_evaluation_examples}
\end{table*}

To parse the answer (in our case, a numerical score) from an answered prompt, we use simple regular expressions to extract the first integer/float score following our input prompt. Table \ref{tab:unparsable_answers} summarizes our findings. For InstructGPT, we rarely see unparsable answers (only a few which are not shown in the table due to rounding). Unparsable answers are far more common for Falcon and Llama2. While the smaller Falcon model with 7b parameters has almost the same fraction of parsable filled prompts as the InstructGPT model, the larger Falcon model with 40b parameters returns many unparsable filled prompts for specific prompts. For each aspect, there are 1-2 prompts where a substantial fraction of the filled prompts are not parsable. For Llama2, we have no clear picture as the 13b shows more parsable filled prompts across most prompts for all evaluation aspects compared to the smaller 7b model. However, the largest 70b model has more unparsable filled prompts than the 13b model.   

We present examples of filled prompts for InstructGPT, Falcon (7b and 40b), and Llama2 (7b, 13b, and 70b) in Table \ref{tab:llm_evaluation_examples} to showcase their characteristics. As can be seen, InstructGPT is usually concise and only returns a score with some explanation. Falcon usually returns a score followed by a more extensive description, while Llama2 is the most verbose, making parsing the actual score challenging since the score might appear at the end of the filled prompt or be outside the maximum sequence length.

Evaluating text style transfer accuracy is more reliable for InstructGPT, Falcon, and Llama2 (except its 70b model) than content preservation. At the same time, fluency is the most reliable (except for the Falcon 40b model).

We can also observe that while for InstructGPT, its only pre-trained counterpart GPT3 shows considerably more unparsable filled prompts, the same does not hold for Falcon and Llama2 (see Appendix \ref{sec:pre_trained_unparsable_filled_prompts} for details). While Falcon's smaller (only pre-trained) 7b model has more unparsable filled prompts than the fine-tuned 7b model, the larger 40b pre-trained model only has more unparsable filled prompts for content preservation. For Llama2, the pre-trained models have more unparsable filled prompts for style transfer accuracy and content preservation for all sizes, while only for fluency, the fine-tuned models have fewer unparsable filled prompts.

\subsection{Out-of-Range Scores}
\begin{table}[htpb!] 
\centering
\begin{tabular}{lrrr}
\hline
 & STA & CP & F \\
\hline
\hline
IGPT & 0.0\% & 0.0\% & 0.0\% \\ 
Falcon-7b & 0.0\% & 0.1\% & 0.3\%\\
Falcon-40b & 0.0\% & 0.0\% & 1.0\% \\
Llama2-7b & 0.4\% & 3.7\% & 0.2\%\\
Llama2-13b & 0.5\% & 0.5\% & 0.2\%\\
Llama2-70b & 0.4\% & 1.0\% & 0.2\%\\
\hline
\end{tabular}
\caption{Shown is the proportion of answers for the three instruction fine-tuned LLM evaluation models InstructGPT (IGPT), Falcon, and Llama2 in different model sizes where the parsed score is out of the given range in the prompt for the aspects of style transfer accuracy (STA), content preservation (CP), and fluency (F). InstructGPT is the most reliable.}
\label{tab:out_of_range_scores}
\end{table}

To count out-of-range scores, we only consider outputs from which we could parse a numerical score. As can be seen from Table \ref{tab:out_of_range_scores}, InstructGPT is again the most reliable, with only a few out-of-range scores. Falcon has slightly more out-of-range scores, and Llama2 has the most out-of-range scores (except for fluency). Apart from content preservation scores for Llama2 with 7b parameters, the scores are usually within the given range in more than 99\% of cases. We can also observe that the LLMs fine-tuned to follow instructions have less or equal out-of-range scores across all settings compared to their pre-trained counterparts (see Appendix \ref{sec:pre_trained_out_of_range_scores} for details).

Out-of-range scores constitute a problem since it is unclear what the upper and lower bounds for potential normalization would be, and outliers can distort the result. We therefore disregard all outputs where scores are out of range. This potentially biases the results if examples are removed that would otherwise have suboptimal scores. Reporting statistics such as inter-annotator agreement, usually done for human evaluations, is difficult as scores can be out of range or non-discrete.

\section{Discussion}
Our results show that the LLMs fine-tuned to follow instructions are a viable tool for TST evaluation. In the case of InstructGPT, we observed that it proves to be a highly effective tool for TST evaluation across various aspects and prompts. Ensembling the scores generated by multiple prompts further enhances the performance of InstructGPT. The LLM evaluations conducted with InstructGPT exhibit a notable correlation with human evaluations, outperforming existing automated metrics in most scenarios related to style transfer accuracy and fluency while remaining on par regarding content preservation. It is worth noting that InstructGPT consistently produces easily interpretable scores, with few out-of-range scores.  

The other two LLMs fine-tuned to follow instructions, Falcon and Llama2, show low or negative correlations with human evaluations for their smaller 7b and 13b parameters. The correlations are often not statistically significant, making them non-reportable due to high p-values. The largest 40b model for Falcon and 70b model for Llama2 also show considerable correlations with human evaluations for all three evaluation aspects. However, they only come close to or surpass InstructGPT for fluency evaluation. The reliability of Falcon and Llama2 in terms of parsable and out-of-range scores is also lower than for InstructGPT. Since Falcon is not fine-tuned with RLHF, we partially attribute its lower performance to this fact. However, one must also consider its smaller size compared to InstructGPT and Llama2. The size is also a factor to be considered when comparing InstructGPT's and Llama2's performance, while we also attribute the more significant variance in Llama2's correlations for TST evaluation to its verbosity.

We also experimented with the not fine-tuned (to follow instructions) versions of InstructGPT (GPT3), Falcon, and Llama2. However, despite generally higher reliability regarding the number of parsable filled prompts, they all showed less correlation with human evaluations, indicating that the returned scores are less reasonable. Since the LLMs fine-tuned to follow instructions share the same architecture as the pre-trained LLMs, we attribute the superior performance to the further alignment with instructions achieved through fine-tuning.

Compared to existing automated metrics, LLM evaluation has the benefit of potentially more explainable results, as already demonstrated by \citet{Cheng:2023}. The prompt can be adapted to ask the LLM to add an explanation to the score. However, as discussed in Section \ref{sec:unparsable_llm_answers}, sometimes these explanations make parsing the results difficult, and there is no guarantee that the explanation matches the returned score. In addition, it is more reproducible than human evaluation, even though the temperature parameter may lead to some variation in outputs. A model can be precisely specified, including its pre-trained weights, random seeds, hyperparameters, and deployed prompts. Therefore, explainability and reproducibility of TST evaluation can be improved using LLMs.

\section{Conclusion \& Future Work}
In this paper, we propose to use LLM evaluation for unified TST evaluation. LLM evaluation can replace existing automated TST evaluation metrics for all three evaluation aspects: style transfer accuracy, content preservation, and fluency. However, we also show that not all LLMs are suitable. LLMs fine-tuned to follow instructions, and especially InstructGPT, seem well-suited. This adds an important puzzle piece to the ongoing efforts to understand capabilities and failure cases of different LLMs.

In the future, we plan to apply our approach to other TST tasks such as formality transfer \cite{Rao:2018} or politeness transfer \cite{Niu:2018}. Multilingual LLMs, such as BLOOM (fine-tuned to follow instructions), also seem promising for a unified multilingual TST evaluation, such as multilingual formality transfer \cite{Briakou:2021b}.

\section*{Limitations}
For our investigation, we had to limit the costs. Therefore, we only considered one particular type of TST, namely sentiment transfer, and the most popular TST dataset, namely Yelp and the human evaluations by \citet{Mir:2019}. To the best of our knowledge and previous studies \cite{Briakou:2021a, Ostheimer:2023}, this is the largest publicly accessible dataset of human evaluations for the monolingual sentiment transfer setting containing outputs from multiple TST models. Limiting the costs also influenced our choice only to use zero-shot prompting. Few-shot prompting would have increased the costs of using GPT3 and InstructGPT using the OpenAI API directly.

On a broader note, high resource usage and costs are inherent LLMs problems. On the one hand, traditional automated evaluation methods such as BLEU \cite{Papineni:2002} for measuring content preservation can be computed within seconds on commodity hardware for a dataset like Yelp. However, more advanced methods that use embeddings, such as WMD \cite{Kusner:2015}, might involve more heavy computations for training the actual embeddings. Furthermore, methods for measuring style transfer accuracy involve training a style classifier such as TextCNN \cite{Kim:2014}, and measuring fluency often involves fine-tuning a language model such as GPT2 \cite{Radford:2019}. On the other hand, prompting one of the largest and most reliable LLMs, such as Falcon or Llama2, results in using multiple GPUs entirely for several hours. LLMs accessible through an API like GPT3 and InstructGPT result in direct costs per submitted token but can also return scores within seconds. However, LLMs do not need any fine-tuning or further training. One major cost driver can here also be prompt engineering \cite{Liu:2023}. To alleviate this issue, we show how prompt ensembling removes the burden of prompt engineering to a certain extent, and just averaging several prompts already results in robust results.

As pointed out by \citet{Ostheimer:2023}, a wide variety of language model architectures and training methods exist to measure fluency in the form of perplexity automatically. The same challenge applies to our method. However, as mentioned, we do not view our method as limited to a particular setup. We demonstrate a unified approach for TST evaluation. Future work on comparing different LLMs is needed.

\section*{Ethical Considerations}
\paragraph{Clarification of the Goals} In addition to the previously discussed limitations of LLM evaluation, a significant ethical concern exists at the core of using LLM evaluation. The question may arise: Is the final goal to replace human evaluation with LLM evaluation? Some may find the idea unsettling, assuming this paper wants to replace humans with LLMs. However, as conscientious and ethical NLP researchers, we want to clarify that this is not our intention. As suggested by our paper's title, we propose an alternative option to unify automated evaluation to enhance the reproducibility and transparency of NLP research. 

\paragraph{Human Evaluations and Experiments} The human evaluations that are used in this paper are provided by \citet{Mir:2019}. We refer to their description of human evaluations. 
Throughout our experiments, we employ models and datasets strictly within their intended usage, ensuring compliance with ethical protocols. Specifically, when using GPT3 and InstructGPT, we adhere to the \href{https://openai.com/policies/usage-policies}{OpenAI usage policy}.
By maintaining a commitment to ethical considerations, we aim to uphold the integrity of our research and contribute to the responsible development and evaluation of AI systems.


\bibliographystyle{acl_natbib}
\bibliography{custom}

\appendix

\section{Largest Pre-trained Models}
This section presents the results for the largest pre-trained LLMs deployed for TST evaluation.

\subsection{Ensembled Prompts vs Other Automated Evaluations}
\begin{table}[ht!]
\centering
\begin{tabular}{lrrrr}
\hline
\multicolumn{5}{c}{Style Transfer Accuracy} \\
\hline
\hline
& ARAE & CAAE & DAR & All\\
\hline
OPT-175b &  \emph{-0.112} & \emph{-0.052} & \emph{0.007} & \emph{-0.039} \\
BLO-176b & 0.311 & \emph{-0.052} & \emph{0.107} & 0.118\\
GPT-175b & 0.126 & \emph{-0.042} & \emph{0.046} & \emph{0.044} \\
Fal-7b & \emph{0.049} & \emph{-0.058} & \emph{0.030} & \emph{0.013} \\
Fal-40b & \emph{0.058} & \emph{0.016} & 0.186 & 0.094\\
Lla-7b & 0.144 & \emph{-0.026} & \emph{-0.014} & \emph{0.030} \\
Lla-13b & 0.271 & \emph{0.109} & 0.215 & 0.191 \\
Lla-70b & 0.350 & 0.406 & 0.389 & 0.393 \\
\hline
\multicolumn{5}{c}{Content Preservation} \\
\hline
\hline
 & ARAE & CAAE & DAR & All \\
\hline
OPT-175b &  \emph{-0.013} & \emph{-0.085} & \emph{-0.080} & \emph{-0.067} \\
BLO-176b & \emph{-0.019} & \emph{-0.051} & \emph{-0.050} & \emph{-0.042}\\
GPT-175b & \emph{0.028} & \emph{0.008} & \emph{0.090} & \emph{0.042} \\
Fal-7b & \emph{-0.070} & \emph{-0.026} & \emph{-0.083} & \emph{-0.036} \\
Fal-40b & \emph{0.047} & \emph{0.104} &  0.189 & 0.114\\
Lla-7b & \emph{-0.009} & \emph{0.000} & \emph{-0.017} & \emph{-0.011} \\
Lla-13b & \emph{0.002} & \emph{0.027} & \emph{0.055} & \emph{0.051} \\
Lla-70b & \emph{0.083} & 0.464 & \emph{0.025} & 0.212 \\
\hline
\multicolumn{5}{c}{Fluency} \\
\hline
\hline
 & ARAE & CAAE & DAR & All \\
\hline
OPT-175b &  \emph{-0.034} & \emph{-0.102} & \emph{0.005} & \emph{-0.058} \\
BLO-176b & -0.127 & \emph{-0.101} & \emph{-0.044} & -0.101\\
GPT-175b & \emph{0.030} & \emph{0.053} & \emph{-0.015} & \emph{0.030}\\
Fal-7b & \emph{-0.019} & \emph{0.079} & \emph{-0.034} & \emph{0.016} \\
Fal-40b & 0.218 & 0.170 & 0.194 & 0.200\\
Lla-7b & \emph{0.107} & 0.199 & \emph{-0.011} & 0.093 \\
Lla-13b & 0.242 & 0.215 & 0.129 & 0.205 \\
Lla-70b & 0.436 & 0.540 & 0.479 & 0.521 \\
\hline
\end{tabular}
\caption{Shown are the Spearman rank correlations for style transfer accuracy (top), content preservation (middle), and fluency (bottom) between human evaluations and the mentioned automated metrics, including OPT, BLOOM (BLO), GPT3 (GPT), Falcon (Fal), and Llama2 (Lla). All \emph{italic correlations} have p>0.05.}
\label{tab:automated_evaluation_correlations_non_instruction}
\end{table}

\subsection{LLM Limitations and Failure Modes}
\subsubsection{Unparsable Filled Prompts}
\label{sec:pre_trained_unparsable_filled_prompts}
Table \ref{tab:pre_trained_unparsable_answers} summarizes our findings in terms of unparsable answers of the pre-trained LLMs. Unparsable answers are common for OPT and BLOOM, where 6.8-14.2\% of the answers of BLOOM and about 6.4-19.2\% OPT's answers are unparsable. GPT3 has considerably lower rates of unparsable answers. We can observe that adding the phrase ``Result ='' as a suffix of the prefix prompt increases the number of parsable answers for OPT, BLOOM, and GPT3. Overall, we can see that evaluating text style transfer accuracy is more reliable for OPT and BLOOM than content preservation and fluency. GPT3 has the worst parsing rates for content preservation.
As shown, the largest investigated language models return a numerical score at least 80\% of the time. To see the effect of model size, we investigate all available pre-trained model sizes of OPT in Appendix \ref{sec:smaller_llms}, demonstrating the increased reliability of bigger models in parsable answers.

\begin{table}[htpb!] 
\centering
\begin{tabular}{lrrr}
\hline
 & STA & CP & F \\
\hline
\hline
OPT & 6.4\% & 11.7\% & 19.2\% \\
BLOOM & 6.8\% & 10.1\% & 14.2\% \\
GPT3 & 3.7\% & 5.8\% & 3.2\% \\
Falcon-7b & 4.7\% & 8.8\% & 2.7\%\\
Falcon-40b & 4.1\% & 10.0\% & 3.5\% \\
Llama2-7b & 3.2\% & 6.0\% & 5.4\%\\
Llama2-13b & 3.1\% & 6.6\% & 4.2\%\\
Llama2-70b & 1.9\% & 4.5\% & 1.9\%\\
\hline
\end{tabular}
\caption{Shown is the proportion of filled prompts for the three largest pre-trained LLM evaluation models OPT, BLOOM, and GPT3 where the answer is not parsable to return a score for the aspects of style transfer accuracy (STA), content preservation (CP), and fluency (F).}
\label{tab:pre_trained_unparsable_answers}
\end{table}

\subsubsection{Out-of-Range Scores}
\label{sec:pre_trained_out_of_range_scores}
As can be seen from Table \ref{tab:pre_trained_out_of_range_scores}, BLOOM has the lowest number of out-of-range scores of the pre-trained LLMs, and OPT has the most out-of-range scores. In contrast to content preservation and fluency evaluations, we observe the least out-of-range scores for style transfer accuracy. We study the effect of the LLM size on the number of out-of-range scores for all available pre-trained OPT models in Appendix \ref{sec:smaller_llms}, showing more out-of-range scores for smaller models.

\begin{table}[htpb!] 
\centering
\begin{tabular}{lrrr}
\hline
 & STA & CP & F \\
\hline
\hline
OPT & 5.2\% & 5.9\% & 6.6\% \\
BLOOM & 1.2\% & 1.3\% & 2.9\% \\
GPT3 & 4.3\% & 2.6 \% & 3.0\% \\
Falcon-7b & 11.2\% & 7,9\% & 1.0\%\\
Falcon-40b & 1.0\% & 3.6\% & 1.1\%\\
Llama2-7b & 3.8\% & 3.4\% & 1.6\%\\
Llama2-13b & 1.2\% & 1.2\% & 0.8\%\\
Llama2-70b & 0.9\% & 1.5\% & 0.8\%\\
\hline
\end{tabular}
\caption{Shown is the proportion of answers for the three largest pre-trained LLM evaluation models OPT, BLOOM, and GPT3 where the parsed score is out of the given range in the prompt for the aspects of style transfer accuracy (STA), content preservation (CP), and fluency (F).}
\label{tab:pre_trained_out_of_range_scores}
\end{table}

\section{Smaller LLM Results for OPT}
\label{sec:smaller_llms}
This section presents the results for smaller LLM evaluations with different OPT sizes.

\subsection{Correlations with Human Evaluations}
As highlighted in Section \ref{sec:effect_of_ensembling}, ensembling enhances the robustness of our LLM evaluation. Therefore, we exclusively report ensembled correlations in this section. We summarize our findings regarding the correlations of smaller LLM evaluations with human evaluations in Table \ref{tab:smaller_llms_correlations} for style transfer accuracy at the top, content preservation in the middle, and fluency at the bottom.

\begin{table}[ht!]
\centering
\begin{tabular}{lrrrr}
\hline
\multicolumn{5}{c}{Style Transfer Accuracy} \\
\hline
\hline
& ARAE & CAAE & DAR & All\\
\hline
OPT125m & \emph{-0.060} & \emph{-0.049} & \emph{0.100} &\emph{0.004}\\
OPT350m & \emph{0.076} & \emph{0.121} & \emph{0.008} &0.074\\
OPT1.3b  & \emph{-0.009} & \emph{0.085} & \emph{0.047} &\emph{0.064}\\
OPT2.7b & \emph{-0.038} & \emph{0.061} & \emph{-0.031} &\emph{-0.000}\\
OPT6.7b & \emph{0.015} & \emph{0.039} & \emph{0.008} &\emph{0.035}\\
OPT13b & \emph{0.018} & \emph{0.011} & \emph{-0.097} &\emph{-0.017}\\
OPT30b & \emph{0.094} & \emph{-0.093} & \emph{-0.079} &\emph{-0.029}\\
OPT66b & \emph{0.051} & \emph{-0.060} & \emph{-0.037} &\emph{-0.016}\\
OPT175b &  \emph{-0.112} & \emph{-0.052} & \emph{0.007} & \emph{-0.039} \\
\hline
\multicolumn{5}{c}{Content Preservation} \\
\hline
\hline
& ARAE & CAAE & DAR & All\\
\hline
OPT125m & \emph{-0.076} & \emph{-0.091} & \emph{-0.028} &\emph{-0.056}\\
OPT350m & \emph{-0.092} & \emph{-0.040} & \emph{-0.036} &\emph{-0.047}\\
OPT1.3b & \emph{0.052} & \emph{-0.010} & \emph{-0.014} &\emph{0.006}\\
OPT2.7b & \emph{0.043} & \emph{-0.002} & \emph{-0.006} &\emph{0.006}\\
OPT6.7b & \emph{-0.076} & \emph{-0.077} & -0.129 &-0.091\\
OPT13b & \emph{-0.012} & \emph{-0.005} & \emph{0.019} &\emph{0.002}\\
OPT30b & \emph{-0.049} & \emph{0.021} & \emph{0.060} &\emph{-0.022}\\
OPT66b & \emph{0.000} & -0.162 & \emph{0.107} &\emph{-0.026}\\
OPT175b &  \emph{-0.013} & \emph{-0.085} & \emph{-0.080} & \emph{-0.067} \\
\hline
\multicolumn{5}{c}{Fluency} \\
\hline
\hline
& ARAE & CAAE & DAR & All\\
\hline
OPT125m & \emph{-0.020} & \emph{-0.029} & \emph{-0.079} &\emph{0.008}\\
OPT350m & \emph{0.064} & \emph{0.068} & \emph{0.023} &\emph{0.051}\\
OPT1.3b & \emph{-0.006} & \emph{-0.054} & -0.133 &-0.078\\
OPT2.7b & \emph{0.029} & \emph{0.031} & \emph{-0.018} &\emph{0.019}\\
OPT6.7b & \emph{-0.034} & \emph{-0.077} & \emph{-0.039} &\emph{-0.069}\\
OPT13b & \emph{-0.067} & \emph{-0.109} & \emph{-0.055} &\emph{-0.084}\\
OPT30b & \emph{-0.111} & -0.165 & \emph{-0.053} &-0.141\\
OPT66b & 0.176 & \emph{0.061} & \emph{0.114} &0.112\\
OPT175b &  \emph{-0.034} & \emph{-0.102} & \emph{0.005} & \emph{-0.058} \\
\hline
\end{tabular}
\caption{Shown are the Spearman rank correlations for style transfer accuracy (top), content preservation (middle), and fluency (bottom) between human evaluations and the LLM evaluations with different model sizes of OPT. All \emph{italic correlations} have p>0.05.}
\label{tab:smaller_llms_correlations}
\end{table}

As mentioned earlier in Section \ref{sec:correlations}, even the largest OPT model with 175 billion parameters exhibits correlations close to zero or slightly negative, with p-values > 0.05 indicating non-reportable correlations for all three evaluation aspects and all investigated TST models (including the combination of their outputs). These results also extend to the smaller LLMs: we also observe correlations close to zero or slightly negative, with p-values > 0.05 for most of the reported correlations on all three evaluation aspects across all investigated TST models.

\subsection{Unparsable Answers}
\begin{figure}[ht!]
\centering
\includegraphics[width=0.45\textwidth]{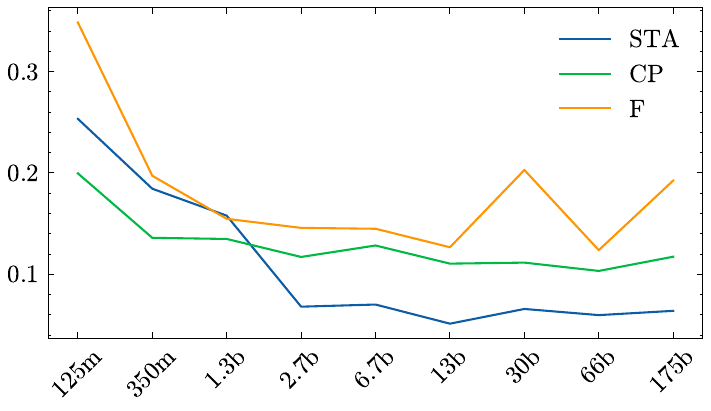}
\caption{Shown is the fraction of unparsable filled prompts for different model sizes of OPT. There is a clear trend that larger models return more parsable filled prompts for all three evaluation aspects of style transfer accuracy (STA), content preservation (CP), and fluency (F).}
\label{fig:smaller_llms_unparsable}
\end{figure}

We summarize our findings for unparsable filled prompts of smaller LLMs in Figure \ref{fig:smaller_llms_unparsable}. Overall, we observe a clear trend: the larger the language model, the more parsable the answers.

For evaluating style transfer accuracy, the fraction of unparsable answers decreases from approximately 0.25 for the smallest 125m model to 0.05 for the 2.7b model and remains at that level. Our analysis shows that bigger models are not necessarily more reliable.

The evaluation of content preservation exhibits a similar trend to style transfer accuracy, with the fraction of unparsable answers decreasing from 0.2 for the smallest 125m model to around 0.12 for the 2.7b model and remaining stable at that level.

For fluency, the fraction of unparsable answers is highest across almost all model sizes (except for the 1.3b model). The fraction of unparsable answers starts at around 0.35 for the smallest 125m model and decreases to approximately 0.25 for the largest models, although the trend is not consistent for the 30b to 175b models. 

\subsection{Out-of-Range Scores}
\label{sec:smaller_llms_oor}
\begin{figure}[ht!]
\centering
\includegraphics[width=0.45\textwidth]{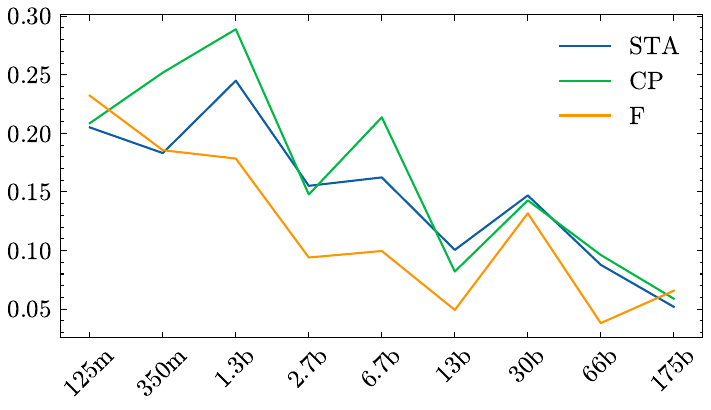}
\caption{Shown is the fraction of filled prompts where the parsed score is out-of-range for different model sizes of OPT. There is a clear trend that larger models return more in-range scores for all three evaluation aspects of style transfer accuracy (STA), content preservation (CP), and fluency (F).}
\label{fig:smaller_llms_oor}
\end{figure}

We summarize our findings for out-of-range scores of smaller LLMs in Figure \ref{fig:smaller_llms_oor}. We observe a similar trend for out-of-range scores as for non-parsable answers: smaller LLMs are less reliable and return more out-of-range scores than larger LLMs. However, the trend exhibits more oscillation compared to non-parsable answers.

In evaluating style transfer accuracy, the fraction of out-of-range scores decreases from approximately 0.2 for the smallest 125m model to 0.05 for the largest 175b model. Intermediate model sizes exhibit oscillation around this downward trend.

Content preservation evaluation follows a similar trend as style transfer accuracy, with the fraction of out-of-range scores decreasing from 0.2 for the smallest 125m model to around 0.05 for the largest 175b model. The oscillation is more pronounced compared to style transfer accuracy.

Fluency evaluation shows the least out-of-range scores starting from the 350m model. The trend remains consistent with the other evaluation aspects: larger models are more reliable than smaller models. The oscillation is smaller, except for the 30b model.

\clearpage

\section{Prompts}
\label{sec:prompts}
We present all deployed prompts for style transfer accuracy in Table \ref{tab:sta_prompts}, for content preservation in Table \ref{tab:cp_prompts}, and for fluency in Table \ref{tab:f_prompts}.

\begin{table*}[ht!]
\centering
\begin{tabular}{lp{13cm}}
\hline
Index & Prompt\\
\hline
\hline
0 & Here is sentence S1: \textbf{\{input\}} and sentence S2: \textbf{\{transferred\}}. How different is sentence S2 compared to S1 on a scale from 1 (identical styles) to 5 (completely different styles)? Result =\\
1 & Here is sentence S1: \textbf{\{input\}} and sentence S2: \textbf{\{transferred\}}. How different is sentence S2 compared to S1 on a continuous scale from 0 (identical styles) to 1 (completely different styles)? Result =\\
2 & Please evaluate the style transfer intensity between sentence A \textbf{\{input\}} and sentence B \textbf{\{transferred\}} on a scale of 1 to 5, where 1 represents an identical style and 5 represents a completely different style.\\
3 & How different is sentence S1 = \textbf{\{input\}} compared to S2 = \textbf{\{transferred\}} on a scale from 1 (identical styles) to 5 (completely different styles)? Result =\\
4 & How different is the sentence S1 = \textbf{\{input\}} compared to S2 = \textbf{\{transferred\}} for style [positivity] on a scale from 1 (identical styles) to 5 (completely different styles)? Result = \\
5 & The sentence S2 = \textbf{\{transferred\}} is a style transfer of sentence S1 = \textbf{\{input\}}, on a scale from 1 (identical styles) to 5 (completely different styles) evaluate the style transfer intensity between S1 and S2? Result =\\
6 & Here is sentence S1: \textbf{\{input\}}, sentence S2: \textbf{\{transferred\}} and style S3 [sentiment]. How different are S1 and S2 for S3 style on a scale from 1 (identical styles) to 5 (completely different styles)? Result =\\
7 &  Here is sentence S1: \textbf{\{input\}}, sentence S2: \textbf{\{transferred\}} and style S3 [sentiment]. How different are S1 and S2 for S3 style on a discrete scale from 1 to 5 where [1 = completely identical styles, 2 =  identical styles, 3 =  not identical nor different styles, 4 = different styles, 5 = completely different styles]? Result =\\
8 & Here is sentence S1: \textbf{\{input\}} and sentence S2: \textbf{\{transferred\}}. How different is sentence S2 compared to S1 on a discrete scale from 1 to 5 where [1 = completely identical styles, 2 =  identical styles, 3 =  not identical nor different styles, 4 = different styles, 5 = completely different styles]? Result =\\
9 & Here is sentence S1: \textbf{\{input\}} and sentence S2: \textbf{\{transferred\}}. How different is sentence S2 compared to S1 on a continuous scale from 1 (completely identical styles) to 5 (completely different styles)? Result =\\
10 & How different is the style of sentence S1 = \textbf{\{input\}} compared to S2 = \textbf{\{transferred\}} on a scale from 1 (identical styles) to 5 (completely different styles)? Result =\\
\hline
\end{tabular}
\caption{Shown are the prompts to measure style transfer accuracy.}
\label{tab:sta_prompts}
\end{table*}

\begin{table*}[ht!]
\centering
\begin{tabular}{lp{13cm}}
\hline
Index & Prompt\\
\hline
\hline
0 & Here is sentence S1: \textbf{\{input\}} and sentence S2: \textbf{\{transferred\}}. The sentences S1 and S2 have the opposite sentiment but how much does the content change on a scale from 1 (completely different content) to 5 (identical content) on a continuous scale? Result =\\
1 & Here is sentence S1: \textbf{\{input\}} and sentence S2: \textbf{\{transferred\}}. The sentences S1 and S2 have the opposite sentiment but has the content changed on a scale from 1 (completely changed) to 5 (not changed)? Result =\\
2 & Here is sentence S1: \textbf{\{input\}} and sentence S2: \textbf{\{transferred\}}. How different is the topic of sentence S2 compared to S1 on a continuous scale from 1 (completely different topic) to 5 (identical topic)? Result =\\
3 & Please rate the content preservation between the following two sentences on a scale from 1 to 5, ignoring any differences in style or formatting: Sentence 1: \textbf{\{input\}} Sentence 2: \textbf{\{transferred\}} To determine the content preservation between these two sentences, consider only the information conveyed by the sentences and ignore any differences in style or formatting. Based on your evaluation, please provide a rating on a scale from 1 to 5, with 1 being very low content preservation and 5 being very high content preservation.\\
4 & Please evaluate the content preservation between sentence A \textbf{\{input\}} and sentence B \textbf{\{transferred\}} on a scale of 1 to 5, where 1 represents identical content and 5 represents completely different content.\\
5 & How much is the content of sentence S2 \textbf{\{input\}} changed from S1 \textbf{\{transferred\}} on a scale from 1 (completely different content) to 5 (identical content)? Result =\\
6 & How much is the content of sentence S2 \textbf{\{input\}} changed from S1 \textbf{\{transferred\}} neglecting the change in style on a scale from 1 (completely different content) to 5 (identical content)? Result =\\
7 & Here is sentence S1: \textbf{\{input\}} and sentence S2: \textbf{\{transferred\}}. The sentences S1 and S2 have the opposite [sentiment], but has the content changed on a scale from 1 to 5 where 5 (Equivalent), 4 (Roughly equivalent), 3 (Not equivalent but share some details), 2 (Not equivalent but on the same topic) and 1 (Completely dissimilar)? Result =\\
8 & Here is the input sentence S1: \textbf{\{input\}} and transferred sentence S2: \textbf{\{transferred\}}. How much does S2 preserve the content of S2 on a scale from 1 to 3 where  3 (The transferred sentence has the same meaning as the input sentence), 2 (The transferred sentence preserves part of the content meaning of the input sentence), and  1 (The transferred sentence and the input sentence are irrelevant to the content)? Result = \\
9 & Here is the input sentence S1: \textbf{\{input\}} and transferred sentence S2: \textbf{\{transferred\}}. How much does S2 preserve the content of S2 on a scale from 1 (completely changed) to 5 (not changed)? Result =\\
10 & Here is S1: \textbf{\{input\}} and sentence S2: \textbf{\{transferred\}}. How much does S2 preserve the content of S2 on a continuous scale from 0 (completely different topic) to 1 (identical topic)? Result =\\
\hline
\end{tabular}
\caption{Shown are the prompts to measure content preservation.}
\label{tab:cp_prompts}
\end{table*}

\begin{table*}[ht!]
\centering
\begin{tabular}{lp{13cm}}
\hline
Index & Prompt\\
\hline
\hline
0 & Here is sentence S1: \textbf{\{input\}} How fluent is the sentence S1 on a continuous scale from 1 (lowest fluency) to 5 (highest fluency)? Result =\\
1 & Please evaluate the fluency, grammar, and naturalness of the following sentence \textbf{\{input\}} on a scale of 1 to 5, where 1 represents poor fluency, grammar, and naturalness and 5 represents excellent fluency, grammar, and naturalness.\\
2 & How fluent is the sentence \textbf{\{input\}} on a continuous scale from 1 (lowest fluency) to 5 (highest fluency)? Result =\\
3 & How fluent is the sentence \textbf{\{input\}} on a continuous scale from 0 (lowest fluency) to 1 (highest fluency)? Result =\\
4 & How fluent is the sentence \textbf{\{input\}} on a scale from 0 to 1 where 0 equals (lowest fluency) and 1 equals (highest fluency)? Result =\\
5 & How fluent is the sentence \textbf{\{input\}} on a scale from 1 to 5 where 5 (perfect), 4 (comprehensible), 3 (somewhat comprehensible), 2 (incomprehensible), and 1 (incomplete or just a fragment)? Result =\\
6 & How are the grammars of the sentence \textbf{\{input\}} on a scale from 1 to 5 where 1 (lowest grammar preserving) and 5 (highest grammar preserving)? Result =\\
7 & On a scale from 1 to 5, rate the fluency and naturalness of sentence S1 \textbf{\{input\}} where 1 (lowest rate) and 5 (highest rate)? Result =\\
8 & On a scale from 1 to 5, how coherent is \textbf{\{input\}} where 1 (lowest coherent) and 5 (highest coherent)? Result =\\
9 & How natural is this sentence S1 \textbf{\{input\}} on a scale from 1 to 5 where 1 (lowest coherent) and 5 (highest coherent)? Result =\\
10 & S1 = \textbf{\{input\}} Rate the fluency of S1 on a scale from 1 (lowest fluency) to 5 (highest fluency).\\
\hline
\end{tabular}
\caption{Shown are the prompts to measure fluency.}
\label{tab:f_prompts}
\end{table*}

\end{document}